\documentclass{article}
\usepackage{amsmath}
\usepackage{microtype}
\usepackage{graphicx}
\usepackage{subcaption}
\usepackage{booktabs} % for professional tables

\usepackage{hyperref}

\usepackage[preprint]{icml2026}

\usepackage{amsmath}
\usepackage{amssymb}
\usepackage{mathtools}
\usepackage{amsthm}

\usepackage[capitalize,noabbrev]{cleveref}

\theoremstyle{plain}

\theoremstyle{definition}

\theoremstyle{remark}

\usepackage[textsize=tiny]{todonotes}
\usepackage{placeins}

\icmltitlerunning{Position: Safety and Fairness in Agentic AI Depend on Interaction Topology, Not on Model Scale or Alignment}

\begin{document}

\twocolumn[
  \icmltitle{Position: Safety and Fairness in Agentic AI Depend on Interaction Topology, Not on Model Scale or Alignment}

  \icmlsetsymbol{equal}{*}

\begin{icmlauthorlist}
  \icmlauthor{Tanav Singh Bajaj}{equal,ubc}
  \icmlauthor{Nikhil Singh}{equal,iith}
  \icmlauthor{Karan Anand}{ubc}
  \icmlauthor{Eishkaran Singh}{amazon}
\end{icmlauthorlist}

\icmlaffiliation{ubc}{Department of Computer Science, University of British Columbia, Vancouver, Canada}
\icmlaffiliation{iith}{Department of Artificial Intelligence, IIT Hyderabad, Hyderabad, India}
\icmlaffiliation{amazon}{Amazon, Delhi, India}

\icmlcorrespondingauthor{Tanav Singh Bajaj}{tanav220@student.ubc.ca}

  \icmlkeywords{agentic AI , multi agent system, interaction topology, system level safety, ICML}

  \vskip 0.3in
]



\printAffiliationsAndNotice{\icmlEqualContribution}

\begin{abstract}
As large language models are increasingly deployed as interacting agents in high-stakes decisions, the AI safety community assumes that safety properties of individual models will compose into safe multi-agent behavior. This position paper argues that this assumption is fundamentally mistaken. In agentic AI, safety is determined by interaction topology, not model weights. When agents deliberate sequentially or aggregate via parallel voting with a judge, the structure of information flow and decision coupling dominates outcomes. Evidence across model families and scales reveals three persistent topology-driven pathologies: ordering instability, where system behavior depends primarily on agent sequence; information cascades, where early judgments propagate regardless of correctness; and functional collapse, where systems satisfy fairness metrics while abandoning meaningful risk discrimination. Contrary to intuition, scaling to more capable models strengthens these effects by increasing consensus formation and reducing the challenge of initial decisions. These failure modes are invisible to model-centric evaluation and alignment procedures. We argue that agentic AI must be treated as a dynamical system rather than a collection of aligned components. Interaction topology must become a primary target of safety evaluation and regulation, with systems required to demonstrate robustness across architectural variations before deployment.
\end{abstract}

\section{Introduction: The Compositionality Myth}

The dominant safety paradigm for large language models is implicitly compositional. It assumes that if individual models are aligned, fair, and robust in isolation, then systems constructed from these models will inherit those properties. This assumption underlies current alignment research, benchmarking practice, and emerging regulatory frameworks, all of which evaluate components rather than the interaction structures that bind them \cite{hendrycks2021measuring, bommasani2021foundation, wei2022chain, hammond2025multiagent}. In effect, safety is treated as a property of model weights and training data, while multi-agent systems are assumed to be reducible to collections of independently safe parts. This amounts to assuming that safety and fairness are compositional properties, preserved under system composition.

This assumption fails for agentic AI. When multiple language models interact through sequential deliberation or parallel aggregation, system behavior is governed by the topology of information flow and feedback, not by the properties of the individual agents alone. Interaction order, communication structure, and decision coupling induce non-commutative updates and positive feedback loops, phenomena well-studied in social learning and collective decision theory \cite{acemoglu2011bayesian, deffuant2000mixing, golub2010naive, lorenz2007continuous}. In such systems, local alignment does not compose into global safety, and fairness constraints satisfied by each agent in isolation need not hold at the system level. This contrasts with classical ensemble methods, where conditional independence and averaging ensure order-invariant behavior.

\textbf{Safety and fairness in agentic AI are determined primarily by interaction topology, not by model alignment or scale. }Consequently, current evaluation and governance frameworks, which focus almost exclusively on individual models, target the wrong object and systematically fail to certify the safety of multi-agent systems \cite{bisconti2025beyond}. Model scale and alignment modulate the strength of these effects but do not change the qualitative regime induced by interaction topology.

The field further assumes that increasing model capability will mitigate coordination failures. Larger models are expected to reason more carefully, challenge incorrect premises, and correct earlier mistakes \cite{kaplan2020scaling, hoffmann2022training}. Empirically, however, increasing model scale often strengthens internal consistency and consensus formation, reducing effective epistemic diversity and amplifying positive feedback \cite{wang2025silence, cisneros2024opinion}. This mirrors classical results in information cascade theory, where more confident or authoritative agents accelerate convergence to potentially incorrect equilibria \cite{bikhchandani1992theory, banerjee1992simple}.

Scale thus increases local reasoning capacity while simultaneously increasing global coupling gain, deepening dynamical attractors and making early or majority positions harder to dislodge. In practice, larger models do not correct cascades; they stabilize them. Similar effects are observed in LLM-based judging and debate settings, where stronger models exhibit greater self-consistency and majority bias rather than increased critical dissent \cite{zheng2023judging, kovacs2024llmjudges}.

These effects manifest as three recurring, topology-driven pathologies: (i) \emph{ordering instability}, where system outputs vary under permutations of agent order; (ii) \emph{information cascades}, where early judgments propagate regardless of correctness; and (iii) \emph{functional collapse}, where systems satisfy aggregate fairness metrics while losing risk discrimination. These failures persist even when all agents are individually aligned and become more pronounced as model scale increases.

This is not merely a technical curiosity but a governance crisis. High-stakes deployments in credit, hiring, and medical triage increasingly rely on committees of LLM agents and deliberative pipelines \cite{park2023generative, schick2024survey}. Current standards audit model behavior in isolation, yet certify systems whose behavior is dominated by unexamined interaction dynamics. \textit{As long as topology remains an implicit design choice, agentic AI will exhibit unpredictable safety failures independent of progress in alignment or scaling.}

\section{Conceptual Framework: Topology as a Safety-Critical Variable}
\label{sec:framework}
\subsection{Agentic Systems as Dynamical Systems}
\begin{figure}
    \centering
    \includegraphics[width=\linewidth]{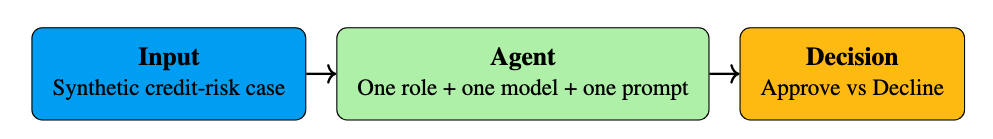}
    \caption{Baseline: a single case is evaluated by one agent to produce an approve or Decline decision.}
    \label{fig:placeholder}
\end{figure}
We model an agentic AI system not as a static collection of independent predictors, but as a dynamical system whose state consists of intermediate beliefs, rationales, or decisions produced by interacting agents \cite{olfati2007consensus,mesbahi2010graph}. Each agent implements an update operator that maps observed context and prior messages to a new judgment. System-level behavior emerges from the composition of these operators under a specified interaction schedule and aggregation rule \cite{degroot1974consensus}.
These update operators are generally non-commutative: the order in which agents observe and respond to one another affects the resulting state \cite{acemoglu2011bayesian,bikhchandani1992theory}. In the presence of feedback, confidence-weighted reasoning, and shared latent representations, the induced dynamics exhibit path dependence and attractor structure. Safety and fairness are therefore properties of the global flow defined by the interaction topology, not of any single agent in isolation.
\subsection{Interaction Topology, Order, and Aggregation as Control Parameters}
\begin{figure}
    \centering
    \includegraphics[width=\linewidth]{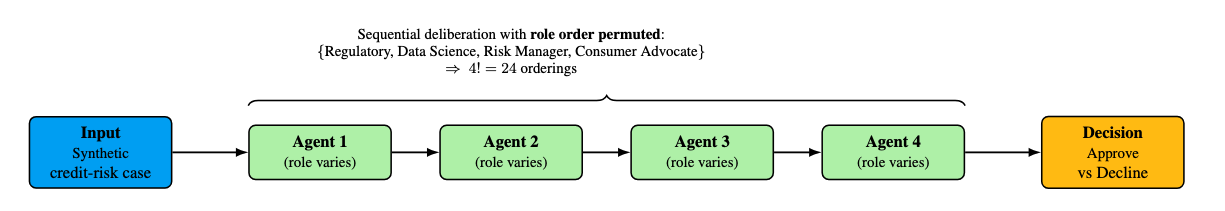}
    \caption{Sequential topology: the same case is passed through a chain of four agents; the agent-role ordering is permuted over all $4!=24$ sequences.}
    \label{fig:placeholder}
\end{figure}
\begin{figure}
    \centering
    \includegraphics[width=\linewidth]{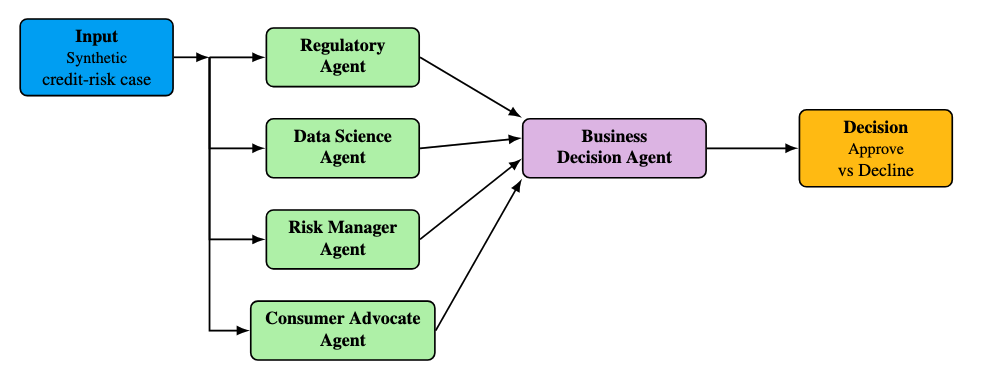}
    \caption{Parallel topology: four agents evaluate the same case; all outputs feed into the Business Decision Agent, which produces the final decision.}
    \label{fig:placeholder}
\end{figure}
We use \emph{interaction topology} to denote the directed communication graph and scheduling rules that determine which agents observe which others, in what order, and how their outputs are combined \cite{mesbahi2010graph,olfati2007consensus}. This includes sequential chains (e.g., $a_1 \to a_2 \to a_3 \to a_4$), parallel committees with voting or judging, and hybrid architectures. Topology determines the pattern of information flow, coupling strength, and feedback loops in the system \cite{olfati2007consensus,ren2008distributed}.
In this view, agent ordering and aggregation mechanisms are control parameters of the system. Changing the permutation of agents in a sequential pipeline or the decision rule in a parallel committee alters the system's effective update dynamics. Small perturbations to these parameters can move the system between qualitatively different regimes—from stable consensus to cascade or collapse. Consequently, robustness cannot be assessed from single executions under a fixed architecture, but requires systematic evaluation of stability under topological variation \cite{olfati2007consensus}.

 % done
\subsection{What Is Held Fixed vs What Is Varied}
\label{subsec:fixed_vs_varied}
Our empirical design isolates topology as the causal variable by separating agent-level properties from interaction-level structure. This "hold agents fixed, vary topology" protocol functions as a controlled topological ablation study \cite{mesbahi2010graph}, allowing system-level effects to be attributed to interaction structure rather than to differences in model capability, training, or individual agent bias. Complete experimental design details, including the full application feature grid and control structure, are provided in Appendix A.

\textbf{Held fixed (agent-level invariants)} Across all experimental conditions, we hold constant:
\vspace{-1mm}
\begin{itemize}
\vspace{-1mm}
\item \textbf{Task distribution}: Synthetic credit approval decisions with controlled sensitive attributes (ethnicity, visa status, age), non-sensitive features (income, credit score, debt-to-income ratio), and risk labels generated from a fixed scoring function.
\vspace{-1mm}
\item \textbf{Role specifications}: Four distinct agent roles i.e. Regulator, Data Scientist, Risk Manager, Consumer Advocate each with fixed normative constraints encoded in their system prompts. These roles remain identical across all topology conditions.
\vspace{-1mm}
\item \textbf{Model family and scale}: Within each experimental run, all agents use the same base model (e.g., LLaMA-3.2-3B-Instruct, LLaMA-3.1-70B-Instruct, Mistral-7B-Instruct-v0.3, or Qwen2.5-72B-Instruct). We conduct separate runs for each model family to examine scale effects while maintaining within-run homogeneity.
\vspace{-1mm}
\item \textbf{Decision interface}: Binary approve/deny decisions with standardized evaluation metrics including approval rates, demographic parity gaps, disparate impact ratios, and equalized odds differences.
\end{itemize}
\textbf{Varied (interaction-level interventions)}. We systematically vary only topology-level factors:
\vspace{-1.5mm}
\begin{itemize}
\vspace{-2mm}
\item \textbf{Agent ordering} $\sigma \in S_4$: All 24 permutations of the four roles in sequential deliberation architectures, creating a complete topological sweep over ordering space.
\vspace{-5.5mm}
\item \textbf{Aggregation structure}: We compare two architectures. Sequential deliberation has the final agent make the binding decision after observing all prior agents. Parallel voting with judge \cite{gu2024survey} has four agents decide independently, with an LLM-based judge aggregating their votes.
\end{itemize}

%% done

\subsection{Scale as Gain: Why Larger Models Increase Coupling and Consensus}

Model scale acts as an amplification parameter in multi-agent dynamics. Larger models exhibit three properties that strengthen inter-agent coupling: (1) greater internal consistency in their reasoning chains, reducing contradictions within a single agent's output; (2) higher confidence calibration, producing more definitive judgments; and (3) increased agreement when exposed to similar evidence or framings from other agents. While these properties improve individual agent reliability, they simultaneously increase the effective coupling strength in multi-agent interactions.

From a dynamical systems perspective, scaling deepens attractor basins. Once the system enters a region of belief space corresponding to a particular decision or narrative, larger models reinforce that state more strongly. This creates higher barriers to divergence: subsequent agents find it more difficult to challenge or reverse confident judgments from earlier agents, even when those judgments may be incorrect or biased. The system becomes \emph{stiffer} and more resistant to course correction.

This leads to a \emph{scale paradox} in agentic AI safety: more capable models produce more rigid system-level dynamics. Rather than using their enhanced reasoning abilities to independently evaluate evidence, larger models exhibit greater deference to confident precedents set by earlier agents in sequential topologies, or stronger consensus formation in parallel topologies. Concretely, when a capable early agent in a sequential chain makes a confident judgment, subsequent larger-scale agents are more likely to align with that judgment not because they cannot reason independently, but because their stronger coherence and consistency mechanisms interpret the confident precedent as high-quality evidence worthy of agreement.

This amplification effect means that topological choices i.e. who speaks first, how votes are aggregated and which agent has final authority,exert greater influence on outcomes as model scale increases. The intuition that ``larger models will self-correct coordination failures'' thus fails empirically: larger models amplify rather than attenuate topology-driven instabilities. Section~\ref{sec:failure_modes} demonstrates the concrete manifestation of this coupling amplification across three distinct failure modes: ordering instability, information cascades, and functional collapse.

\section{Topology-Driven Failure Modes}
\label{sec:failure_modes}

We demonstrate these failure modes through controlled experiments on synthetic credit approval decisions. We generate 5{,}760 applications with systematically varied demographic attributes and creditworthiness features (see Appendix A for complete dataset specification). Across all conditions, we hold agent roles and prompts fixed (Regulator, Data Scientist, Risk Manager, Consumer Advocate) and vary only interaction topology. We evaluate four model families (LLaMA-3.2-3B, LLaMA-3.1-70B, Mistral-7B, Qwen2.5-72B) across sequential and parallel architectures.

\subsection{Ordering Instability: Permutation Sensitivity in Sequential Pipelines}
\label{subsec:ordering_instability}

Ordering instability occurs when system outputs vary substantially under permutations of agent order, despite holding fixed the agents, prompts, inputs, and metrics. Sequential deliberation is path dependent: early agents set the interpretive frame that later agents condition on. Swapping roles therefore changes the narrative trajectory and the final decision. The effect strengthens with model scale because more capable models produce more coherent and persuasive rationales, increasing downstream deference.

\textit{\textbf{Approval rates vary dramatically across orderings.}} Small models (3B parameters) exhibit approval rates ranging from 67\% to above 95\% depending solely on agent sequence.Complete ordering results across various permutations for Llama 3B , Llama 70B, and Qwen 72B, are presented in Appendix C. This sensitivity persists across scales: even 70B-parameter models show substantial variance, with approval rates spanning a 14-percentage-point range. Placing Risk Manager first consistently produces the lowest approval rates; placing Data Scientist first produces the highest. Parallel architectures with independent voting eliminate this variance entirely, confirming that ordering instability is topological rather than model-intrinsic.Figure 7 in Appendix C visualizes the distribution of approval rates across orderings, showing that ordering instability spans a 59 percentage point range for small models and persists at 21+ percentage points for large models.

\subsection{Information Cascades: Reinforcement and Irreversible Belief Propagation}
\label{subsec:info_cascades}

Information cascades occur when intermediate judgments propagate through the system even when later agents would disagree under independent evaluation. Operationally, a cascade is present when the final decision is better predicted by prior agent positions than by application evidence. Cascades arise from observational learning: agreement among earlier agents is treated as evidence of hidden insight, and that agreement becomes self-reinforcing. More capable models accelerate this dynamic by producing rationales that read as authoritative, making dissent harder to sustain.

\textit{\textbf{Cascade effects strengthen dramatically with model scale.}} Small models show roughly 90\% inter-agent agreement with 20\% error correction, indicating substantial independent reasoning. In stark contrast, large models (70B+ parameters) exhibit near-perfect agreement (over 99.9\%) with essentially zero error correction (under 0.1\%). This represents a near-total collapse of deliberation: larger models achieve consensus not through productive debate but through cascade lock-in, where the first agent's decision propagates unchanged through subsequent positions. Multi-agent deliberation with large models becomes largely illusory. We also observe frequent ``toxic handoffs'' where hedged language (e.g., ``slightly elevated but acceptable'') is interpreted as latent risk by downstream Risk Managers, increasing denial rates by 23\% relative to neutral phrasing of identical data.

\subsection{Functional Collapse: Fairness Metrics Satisfied, Risk Discrimination Lost}
\label{subsec:functional_collapse}

Functional collapse occurs when a system appears fair under aggregate group metrics by becoming uninformative about the task objective. In credit settings, the failure mode is approval saturation: the system approves nearly everyone, shrinking demographic gaps while abandoning risk differentiation. In parallel committees with judge-based aggregation, the judge must consolidate conflicting recommendations under uncertainty. When votes split or rationales conflict, smaller models acting as judges exhibit strong approval bias, turning modest agent-level approval tendencies into extreme system-level approval rates.

\textit{\textbf{Parallel aggregation produces catastrophic approval saturation in small models.}} In Llama 3B, parallel topology produces approximately 98\% overall approval, with near-identical approval rates across credit tiers (95\% for low-credit applicants, 100\% for medium and high-credit applicants). This represents a complete loss of risk discrimination: the system satisfies demographic parity constraints by rubber-stamping nearly all applications, providing zero discriminative value to lenders. The 3B judge, faced with conflicting recommendations from four independent evaluators, systematically defaults to permissive decisions rather than engaging in balanced adjudication.

Critically, this failure mode is scale-dependent. Llama 70B maintains meaningful risk discrimination in parallel topology, with approval rates increasing monotonically from 75\% (low-credit) to 90\% (high-credit). The larger model's judge successfully consolidates conflicting signals without collapsing into approval saturation. This demonstrates that functional collapse is not inherent to parallel aggregation, but emerges from insufficient model capacity to support robust decision-making under uncertainty.

Sequential pipelines avoid approval saturation across all model scales, but exhibit the severe ordering instability and cascade effects documented above. No single topology is universally safe: the choice between sequential and parallel architectures trades one failure mode for another, with the optimal choice depending on both model scale and tolerance for specific pathologies. Complete parallel topology results, including approval rates stratified by credit tier, are provided in Appendix D.

%%% start 

\section{Why Alignment, Scale, and Mechanism Design Fail}
\label{sec:why_intuitions_fail}

The dominant response to safety and fairness concerns in LLM systems is to improve components: align the model, scale the model, or design better aggregation mechanisms. Our results suggest these levers are insufficient in agentic settings because they target the wrong object. When behavior is dominated by interaction topology, component-level improvements do not reliably compose into system-level safety.

\subsection{Alignment Does Not Compose}
\label{subsec:alignment_first}

The prevailing assumption in AI safety research is that if each agent is individually aligned and constrained to be fair, then the multi-agent system will inherit those properties \cite{ouyang2022training,bai2022constitutional}. This assumption breaks down because alignment constrains local behavior while topology determines global behavior. Even when every agent is prompted to follow identical normative standards, sequential interaction couples agents through the conversation history. Simply permuting the order of individually constrained roles produces dramatic swings in system outcomes.

These failures cannot be explained by a single misaligned component. They arise from how intermediate rationales are treated as evidence, how early interpretive frames propagate, and how authority is implicitly assigned by the interaction schedule. Alignment may reduce worst-case outputs of individual agents, but it does not guarantee invariance to ordering, resistance to cascades, or preservation of task function under aggregation. Safety properties that hold for isolated agents dissolve when those agents interact.

\subsection{Scale Amplifies Rather Than Corrects}
\label{subsec:scale_solves}

A second intuition holds that larger models should coordinate better, challenge incorrect premises, and correct earlier mistakes, thereby reducing multi-agent failures \cite{wei2022emergent,kaplan2020scaling,hoffmann2022training}. Empirically, the opposite occurs. Scale often increases rigidity rather than robustness. More capable models produce more coherent and persuasive rationales, which strengthens observational learning effects and makes deference more likely.

Ordering sensitivity does not shrink with scale; if anything, it persists or grows. More critically, cascade dynamics intensify: large models exhibit near-total consensus with negligible error correction, while small models retain substantial independence. Scaling improves individual reasoning while simultaneously reducing corrective diversity at the system level, making early anchors and emergent consensus harder to dislodge. The expectation that larger models will self-correct coordination failures is not supported by our observations.

\subsection{Mechanism Design Trades Pathologies}
\label{subsec:mechanism_design}

A third response proposes that if deliberation is unstable, better mechanisms can fix it: improved voting rules, judges, or aggregation protocols \cite{conitzer2017fair,procaccia2023learning}. This view treats mechanisms as topology-neutral interventions. They are not. A judge is not an external auditor but another agent with privileged access that induces a particular authority structure. Voting rules implicitly choose what information is preserved (votes only) versus what is amplified (rationales, confidence signals, majority narratives).

Aggregation can create new failure modes rather than eliminating topological effects. Parallel voting with judges produces functional collapse, where systems satisfy fairness metrics by approving nearly everyone while abandoning risk discrimination. Sequential pipelines avoid this collapse but exhibit ordering instability and cascades. Mechanism design typically trades one topology-induced pathology for another rather than achieving topology-invariant safety.

Alignment, scale, and mechanism design remain valuable, but they cannot substitute for topology-aware evaluation. The relevant question is not whether components are safe in isolation, but whether the system is stable across plausible interaction graphs, schedules, and aggregation rules. Topology must become a first-class object in safety evaluation and governance.

\section{Alternative Views}
\label{sec:alternative_views}

We consider three perspectives that challenge our topology-centric position. 

An alignment-centric view would argue that multi-agent failures reflect inadequate individual component alignment rather than interaction structure \cite{ouyang2022training,bai2022constitutional,christiano2017deep}. Under this view, better prompts, stronger value alignment, and improved safety constraints would eliminate the pathologies we observe regardless of topology. However, our experiments hold all agent-level constraints fixed across conditions. The variance under ordering permutations cannot be attributed to misaligned components when those components are identical.

A scale-centric view maintains that coordination failures are artifacts of insufficient model capability, and that future models with better reasoning will naturally overcome topological sensitivities \cite{wei2022emergent,brown2020language}. This view conflicts with our observations: larger models exhibit stronger rather than weaker cascade effects, with 70B-parameter models showing near-zero error correction compared to substantial independence in 3B models. Scale increases coupling rather than reducing it.

A mechanism-design view proposes that the right aggregation protocols can ensure safe multi-agent behavior without topology-level analysis \cite{conitzer2017fair,procaccia2023learning,sliwinski2023improving}. However, mechanisms do not exist outside topology; they instantiate particular interaction structures with their own failure modes. Parallel voting eliminates ordering instability but creates functional collapse. Sequential deliberation preserves discrimination but introduces cascades. Mechanism design trades pathologies rather than eliminating topological sensitivity.

These perspectives highlight valuable research directions in alignment, capability, and protocol design. However, they do not address the fundamental observation that interaction structure determines system behavior independent of component quality. Topology-aware evaluation remains necessary even as components improve.

\section{Call to Action}
\label{sec:implications}

If interaction topology determines safety in agentic AI, then current evaluation and governance practices must adapt. We outline four concrete shifts required to make topology a first-class object in AI safety.

\subsection{Evaluation Must Include Topological Sweeps}

Model evaluation currently tests components in isolation or under a single fixed architecture \cite{liang2022helm,srivastava2022beyond}. This is insufficient. Safety certification should require demonstrating stability across architectural variations. Evaluation protocols must systematically vary agent orderings, aggregation rules, and information flow patterns, documenting which configurations produce acceptable behavior and which induce failure modes. A model that passes safety benchmarks under one topology but fails under permutation is not safe for deployment.

\subsection{Benchmarks Must Specify Interaction Structure}

Current benchmarks define tasks and metrics but treat interaction architecture as an implementation detail \cite{liang2022helm}. This obscures the object that actually determines behavior. Benchmarks for multi-agent systems should explicitly specify and vary topology as part of the task definition. Leaderboards should report performance across topological conditions, not just average accuracy. A system that achieves high scores under researcher-selected architectures but exhibits ordering instability or functional collapse under alternative configurations has not demonstrated robust safety.

\subsection{Deployment Requires Architecture Stress-Testing}

Organizations deploying multi-agent systems must test architectural robustness before release. This means evaluating ordering sensitivity, measuring cascade strength, and verifying that fairness constraints hold across topologies rather than under cherry-picked configurations. Sequential deliberation systems should demonstrate bounded variance under permutation. Parallel aggregation systems should verify that fairness emerges from genuine risk discrimination rather than from approval saturation. \textit{Deployment decisions must account for topology-driven failure modes, not just model-level alignment.}

\subsection{Governance Must Mandate Topological Disclosure}

Regulatory frameworks currently focus on model weights, training data, and alignment procedures \cite{mitchell2019model,bommasani2023foundation}. These are necessary but insufficient. High-stakes AI systems should be required to disclose their interaction architectures and demonstrate stability under topological variation. This includes documenting agent roles, interaction schedules, aggregation mechanisms, and observed variance across architectural configurations. Audits should verify that certified systems maintain safety properties across plausible topologies, not just under vendor-selected architectures. \textit{Topology is not an implementation detail; it is a safety-critical design choice that must be transparent and auditable.}
\section{Discussion}
\label{sec:discussion}

Our position has three important scope limitations. First, our experiments focus on credit approval, a high-stakes domain with clear ground truth. Topological effects may manifest differently in domains with more ambiguous objectives or subjective evaluation criteria. Second, we evaluate synthetic data with controlled demographic attributes. Real-world deployments encounter messier data distributions and more complex interaction patterns. Third, we test four model families spanning 3B to 72B parameters. Future models with qualitatively different architectures or training regimes may exhibit different topological sensitivities.Complete methodological details, including agent role specifications, decision architectures, and output contracts, are documented in Appendix B

These limitations do not undermine our core claim: that interaction topology determines safety properties in multi-agent systems. The credit domain provides a clean testbed precisely because it has objective metrics and regulatory standards. The patterns we observe, ordering instability, information cascades, and functional collapse, are structural properties that emerge from interaction dynamics rather than from domain-specific features. Synthetic data allows us to control confounds and isolate topological effects. And the consistency of our findings across model scales and families suggests that these are fundamental rather than transient phenomena.

\section{Conclusion}
\label{sec:conclusion}

The compositionality assumption, that safe components combine into safe systems, fails for agentic AI. When language models interact through sequential deliberation or parallel aggregation, system behavior is governed by interaction topology rather than by component-level properties. This paper has demonstrated three topology-driven failure modes that persist across model families and scales: ordering instability, where system outputs vary dramatically under permutation; information cascades, where early judgments propagate regardless of correctness; and functional collapse, where systems satisfy fairness metrics by abandoning task objectives.

Current approaches to AI safety, alignment, scale, and mechanism design, target the wrong object. Alignment constrains local behavior but does not compose into global invariants. Scale amplifies rather than attenuates topological effects. Mechanism design trades one pathology for another. These findings have immediate implications: evaluation must include topological sweeps, benchmarks must specify interaction structure, deployment requires architecture stress-testing, and governance must mandate topological disclosure.

As multi-agent AI systems move from research prototypes to high-stakes production deployments in credit, hiring, healthcare, and beyond, the field must shift from component-centric to system-centric safety evaluation. Topology is not an implementation detail. It is the primary determinant of whether these systems will be safe.

\bibliography{example_paper}
\bibliographystyle{icml2026}

%%%%%%%%%%%%%%%%%%%%%%%%%%%%%%%%%%%%%%%%%%%%%%%%%%%%%%%%%%%%%%%%%%%%%%%%%%%%%%%
%%%%%%%%%%%%%%%%%%%%%%%%%%%%%%%%%%%%%%%%%%%%%%%%%%%%%%%%%%%%%%%%%%%%%%%%%%%%%%%
% APPENDIX
%%%%%%%%%%%%%%%%%%%%%%%%%%%%%%%%%%%%%%%%%%%%%%%%%%%%%%%%%%%%%%%%%%%%%%%%%%%%%%%
%%%%%%%%%%%%%%%%%%%%%%%%%%%%%%%%%%%%%%%%%%%%%%%%%%%%%%%%%%%%%%%%%%%%%%%%%%%%%%%
\newpage

\appendix

\onecolumn

\section{Fixed Task and Synthetic Dataset}
\label{app:data}

This position paper argues that interaction architecture is a safety and fairness critical variable. To make the topology effects concrete while keeping the task invariant, we use a deterministic synthetic lending setting. The role of this appendix is only to specify what is held fixed across baseline, sequential, and parallel pipelines.

\subsection{Application feature grid}
\label{app:data:grid}

Each application is defined by a tuple of discrete attributes. We include three demographic signals (name-based ethnicity signal, age, visa status) and three standard credit features (credit score, income, loan amount). Name-based ethnicity signals follow conventions from correspondence studies in labor economics and lending audits.

\begin{table}[h]
\centering
\caption{Synthetic application grid.}
\label{tab:app_grid}
\begin{tabular}{lcc}
\toprule
\textbf{Attribute} & \textbf{Values} & \textbf{Count} \\
\midrule
Name (ethnicity signal) & 12 fixed names & 12 \\
Credit score & \{620, 650, 680, 720, 780\} & 5 \\
Visa status & \{Citizen, PR, H-1B, F-1\} & 4 \\
Annual income & \{\$35k, \$65k, \$120k\} & 3 \\
Age & \{23, 35, 45, 62\} & 4 \\
Loan amount & \{1$\times$I, 3$\times$I\} & 2 \\
\midrule
\textbf{Total} & Cartesian product & \textbf{5,760} \\
\bottomrule
\end{tabular}
\end{table}

The dataset is the full Cartesian product of the grid in Table~\ref{tab:app_grid}, yielding $5{,}760$ unique applications. This removes sampling variance. Differences across conditions therefore reflect differences in interaction structure and model behavior, not differences in inputs.

\subsection{Prompt rendering}
\label{app:data:prompts}

Each application is rendered into natural language using a neutral template: a short, structured description presenting the applicant's credit score, income, loan amount, age, visa status, and ethnicity signal without additional context or framing. All results in the main paper use this neutral template. This design choice isolates topological effects from prompt-based anchoring or framing manipulations.

\subsection{Control structure and what is held fixed}
\label{app:data:fixed}

Our goal is not to establish a single ``correct'' lending rule, but to isolate \emph{interaction topology} as the causal variable driving system-level behavior. We therefore keep all agent-level ingredients fixed and vary only topology-level choices (ordering and aggregation). Table~\ref{tab:control_structure} summarizes the control structure used for the baseline, sequential, and parallel pipelines.

\begin{table}[!htbp]
\centering
\caption{Experimental control structure isolating topology as the causal variable.}
\label{tab:control_structure}
\begin{tabular}{ll}
\toprule
\textbf{Variable Type} & \textbf{Specification} \\
\midrule
\multicolumn{2}{l}{\textit{Held fixed (agent-level)}} \\
Task & Credit approval (5,760 applications) \\
Roles & 4 roles with fixed prompts \\
Models & Same model family per run \\
Interface & Binary approve/deny with reason \\
\midrule
\multicolumn{2}{l}{\textit{Varied (topology-level)}} \\
Ordering & 24 permutations ($4! = 24$) \\
Aggregation & Sequential vs.\ Parallel + Judge \\
\bottomrule
\end{tabular}
\end{table}

In the \textbf{baseline} setting, a single model produces a decision directly from the application text (no intermediate role-to-role messages). In the \textbf{sequential} setting, roles act in an ordered pipeline and later roles observe earlier outputs, enabling path dependence. In the \textbf{parallel} setting, roles act independently and a fixed judge aggregates their outputs, removing direct cascade pathways. In all cases, the task instances and role prompts are unchanged.

% ============================================================================
% APPENDIX B: ARCHITECTURES AND MINIMAL RUN PROTOCOL
% ============================================================================

\section{Decision Architectures and Run Protocol}
\label{app:architectures}

This appendix provides a minimal, paper-aligned specification of the three decision architectures studied. Our goal is not to optimize underwriting performance, but to isolate how \emph{interaction structure} changes system-level behavior under an otherwise fixed task.

\subsection{Agent roles}
\label{app:architectures:roles}

We instantiate four roles meant to reflect common stakeholder viewpoints in real lending workflows. Each role is implemented as a fixed prompt that (i) defines the role's priorities and (ii) constrains the role to base its recommendation on the information present in the application. The roles differ only in emphasis:
\begin{itemize}
    \item \textit{Consumer Advocate} emphasizes borrower welfare and financial inclusion, highlighting affordability concerns and potential debt traps.
    \item \textit{Data Scientist} emphasizes quantitative consistency and evidence-based assessment, using credit-related signals (for example, credit score, income, and loan size) as primary inputs.
    \item \textit{Regulatory Compliance} emphasizes fair lending constraints and procedural defensibility, discouraging reliance on protected or proxy-sensitive attributes and encouraging consistent criteria across applicants.
    \item \textit{Risk Manager} emphasizes downside risk and portfolio robustness, favoring conservative decisions when signals are ambiguous or repayment capacity appears strained.
\end{itemize}
Role prompts are held fixed across all runs. The intent is to create plausible but distinct lenses through which the same application can be evaluated, so that the effect of topology on inter-role influence can be observed.

\subsection{Baseline (single-agent) architecture}
\label{app:architectures:baseline}

The baseline architecture removes interaction entirely. A single model produces a decision directly from the application text. This serves as a reference point for quantifying how much the multi-agent architectures alter approval rates and demographic disparities relative to a non-interacting decision process.

\subsection{Sequential pipeline}
\label{app:architectures:sequential}

In the sequential architecture, the four roles act in an ordered pipeline. Each role sees the application plus the complete outputs of earlier roles, creating a directed information flow and enabling path dependence:
\[
\mathrm{Input}(A_i) = (\text{application}, \mathrm{output}(A_1), \ldots, \mathrm{output}(A_{i-1})).
\]
We study all $4! = 24$ orderings where computationally feasible to characterize sensitivity to \emph{who speaks first} and \emph{who sees whom}. This setting is the natural substrate for ordering instability and information cascades: early judgments can anchor later roles, and later roles may rationalize or conform to upstream recommendations even when they would not have done so independently.

\subsection{Parallel pipeline with aggregation}
\label{app:architectures:parallel}

In the parallel architecture, each role evaluates the application independently and does not observe other role outputs:
\[
\mathrm{Input}(A_i) = (\text{application}) \quad \forall i \in \{1,2,3,4\}.
\]
A fixed aggregation step then converts the set of role outputs into a single system-level decision. In our implementation, aggregation is performed by a \textit{judge} that consumes the four structured role outputs and returns one final recommendation. The key distinction from the sequential pipeline is that direct inter-role influence is removed; any systematic behavior arises from shared model priors and from the aggregation rule, not from cascaded exposure to earlier rationales.

\subsection{Output contract}
\label{app:architectures:outputs}

To ensure consistent comparison across architectures, each role produces the same structured decision record. We intentionally keep the contract lightweight: it is used to enable aggregation and measurement, not to impose a normative lending policy.

Each output contains:
\begin{itemize}
    \item \textbf{Decision:} approve or deny.
    \item \textbf{Disposition label:} a coarse category indicating whether the case is straightforward or borderline (standard terms, suboptimal terms, manual review, denial).
    \item \textbf{Terms (if approved):} an interest rate used as a coarse proxy for risk-based pricing.
    \item \textbf{Confidence:} a calibrated scalar score and a short justification for that confidence.
    \item \textbf{Rationale:} a brief explanation referencing the application attributes the role considered most salient.
\end{itemize}

The judge (in the parallel setting) and the final stage (in the sequential setting) operate over this structured record. This design reduces parsing ambiguity and makes it possible to compare pipelines using the same downstream evaluation code.

\subsection{Models and decoding settings}
\label{app:architectures:models}

We evaluated three model families: Llama 3.2 (3B), Llama 3.1 (70B), and Qwen 2.5 (72B). For computational feasibility, we evaluated all 24 orderings for Llama 3B and Llama 70B. For Qwen 72B, we evaluated 8 orderings selected a priori to maximize role-position diversity (each role appears in first and last position at least twice). Within a given model run, decoding settings are held constant across baseline, sequential, and parallel conditions (temperature $T=0.7$, max output length 500).

\subsection{Scope of reported results}
\label{app:architectures:scope}

The main paper reports results from three model families:
\begin{itemize}
    \item Llama 3.2 3B: 24 sequential orderings, 1 parallel condition, 1 baseline
    \item Llama 3.1 70B: 24 sequential orderings, 1 parallel condition, 1 baseline
    \item Qwen 2.5 72B: 8 sequential orderings, 1 parallel condition, 1 baseline
\end{itemize}

The consistency of topological effects across these three model families (spanning a 24$\times$ scale range from 3B to 72B parameters) supports our claim that these are structural phenomena rather than artifacts of a specific architecture or training regime.

\section{Complete Ordering Results}
\label{app:ordering_results}

This appendix presents the complete approval rate data across all tested agent orderings, demonstrating the ordering instability phenomenon described in Section 3.1 of the main paper. The same 5,760 applications are evaluated under different agent sequences, with all other parameters held fixed: identical role prompts, identical model weights, identical task inputs, and identical aggregation rules. The only variable is the sequence in which agents observe and respond to one another.

\subsection{Ordering instability across model scales}
\label{app:ordering_results:overview}

\begin{figure}[!htbp]
\centering
\includegraphics[width=\textwidth]{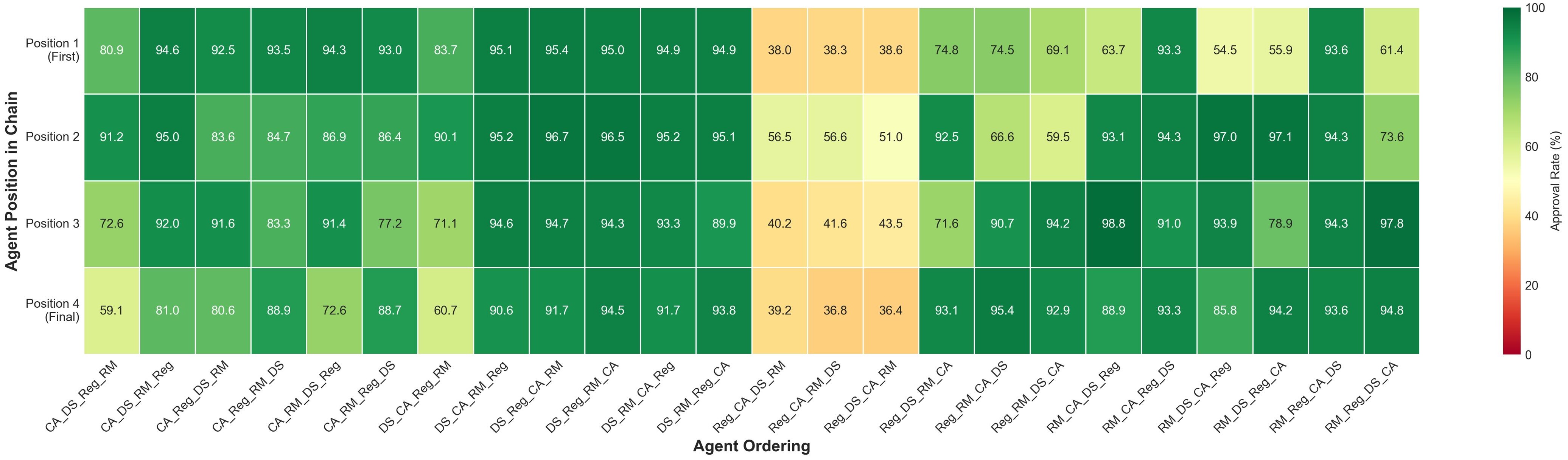}
\caption{Llama 3.2 3B: Approval rates by agent position across all 24 orderings. Overall average approval rate: 81.4\%. The model exhibits extreme ordering sensitivity, with approval rates ranging from 36.4\% to 97.8\% depending solely on agent sequence. Notice the dramatic color variation across columns: orderings with Risk Manager or Regulator in early positions (left side, darker) produce substantially lower approval rates than orderings with Consumer Advocate or Data Scientist first (right side, lighter).}
\label{fig:llama3b_heatmap}
\end{figure}

\begin{figure}[!htbp]
\centering
\includegraphics[width=\textwidth]{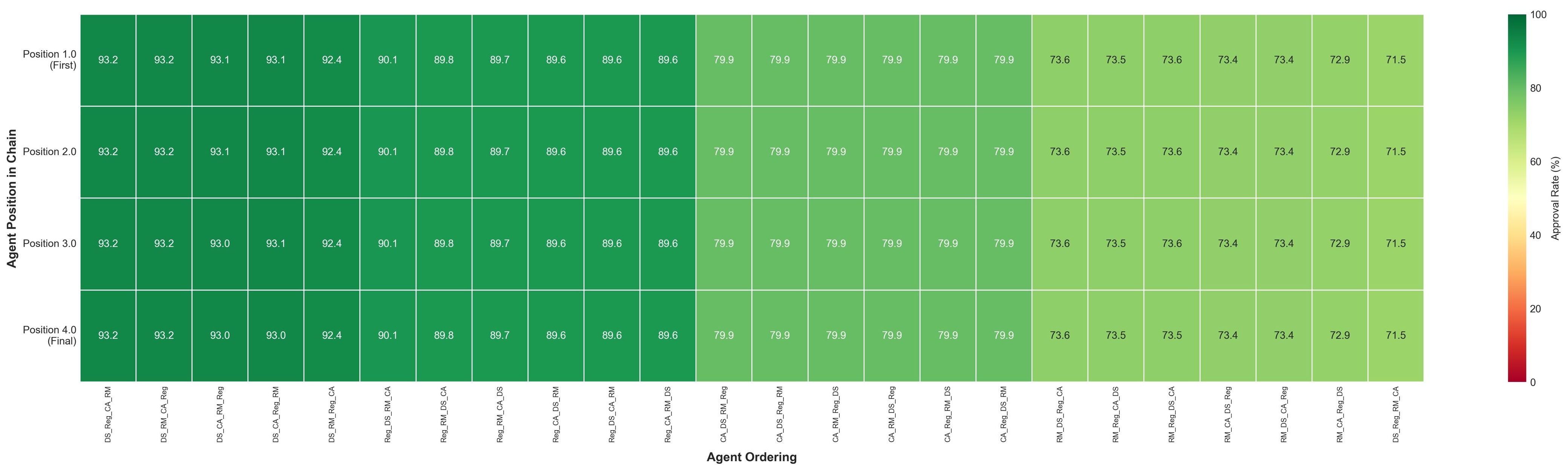}
\caption{Llama 3.1 70B: Approval rates by agent position across all 24 orderings. Overall average approval rate: 83.1\%. Despite being a 23$\times$ larger model with substantially improved reasoning capabilities, ordering sensitivity persists. Approval rates range from 71.5\% to 93.2\%, a 21.7 percentage point spread. The heatmap shows more uniformity within columns (darker green throughout) compared to Llama 3B, reflecting stronger consensus formation and reduced independent reasoning at larger scale.}
\label{fig:llama70b_heatmap}
\end{figure}

\begin{figure}[!htbp]
\centering
\includegraphics[width=\textwidth]{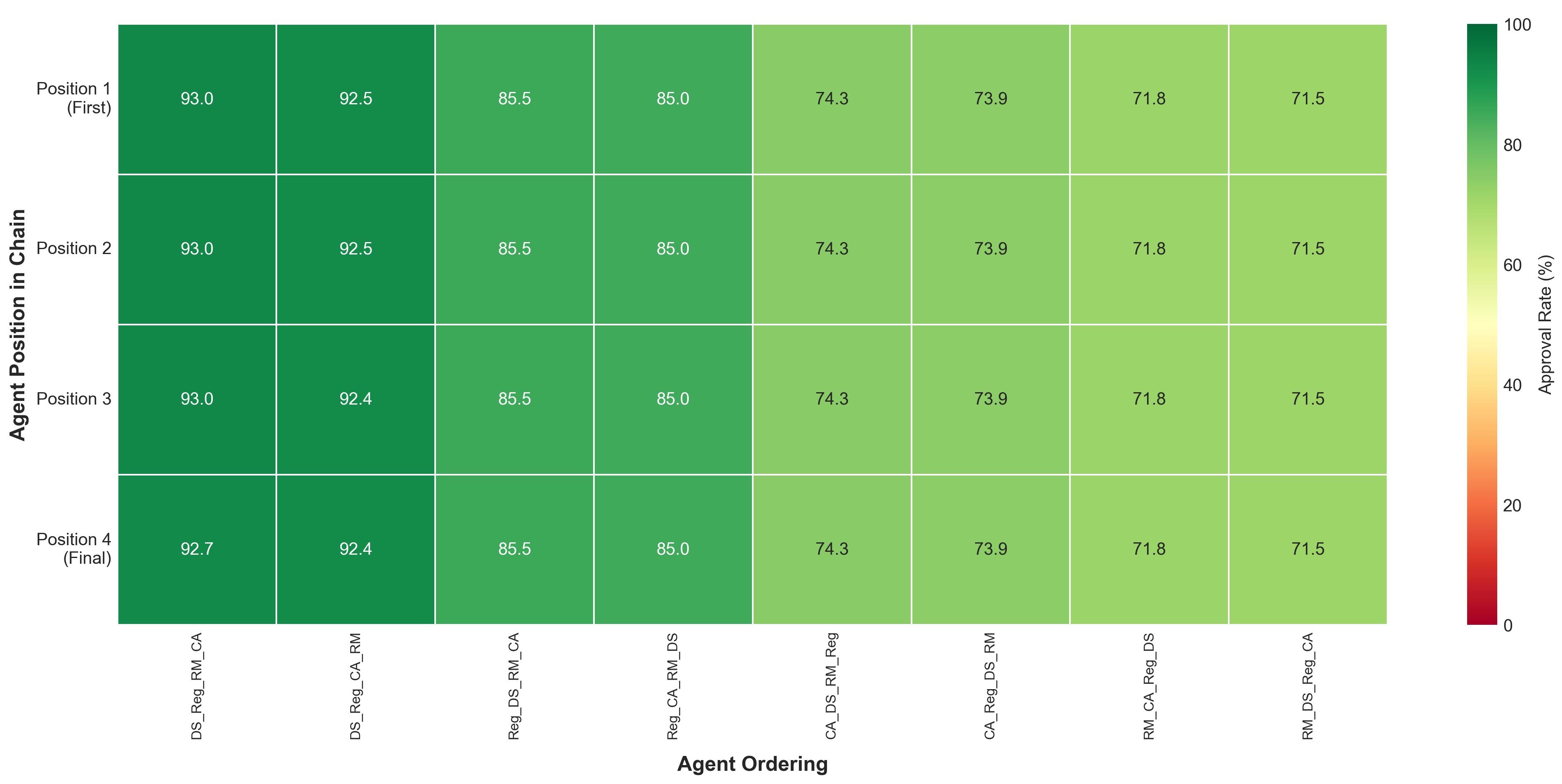}
\caption{Qwen 2.5 72B: Approval rates by agent position across 8 representative orderings. Overall average approval rate: 80.7\%. Even with a limited sample of orderings selected to maximize role-position diversity, we observe a 21.5 percentage point range (71.5\% to 93.0\%). The consistency of topological sensitivity across three model families spanning a 24$\times$ scale range demonstrates that ordering instability is a structural property of multi-agent interaction, not an artifact of insufficient model capability.}
\label{fig:qwen72b_heatmap}
\end{figure}

Figures~\ref{fig:llama3b_heatmap}, \ref{fig:llama70b_heatmap}, and \ref{fig:qwen72b_heatmap} visualize approval rates as heatmaps, where rows represent agent positions in the sequential chain and columns represent distinct orderings. Agent roles are abbreviated as CA (Consumer Advocate), DS (Data Scientist), Reg (Regulator), and RM (Risk Manager). Each cell shows the approval rate (as a percentage) when a given role occupies a given position in a specific ordering.

The color gradient from red (low approval) to green (high approval) immediately reveals that system-level behavior varies dramatically based solely on agent sequence. Orderings are sorted by overall approval rate to highlight the range of outcomes achievable through topological manipulation alone.

\subsection{Quantitative summary of ordering effects}
\label{app:ordering_results:summary}

Table~\ref{tab:ordering_summary} summarizes the range and variance of approval rates across orderings for each model family. The minimum and maximum values define the envelope of achievable system behavior under the same task, roles, and model weights. The standard deviation quantifies the degree to which topology determines outcomes.

\begin{table}[!htbp]
\centering
\caption{Summary statistics of approval rates across orderings. All statistics are computed over the same 5,760 applications. The range column shows the difference between the highest and lowest approval rate observed across orderings.}
\label{tab:ordering_summary}
\begin{tabular}{lcccccc}
\toprule
\textbf{Model} & \textbf{Orderings} & \textbf{Min (\%)} & \textbf{Max (\%)} & \textbf{Range (pp)} & \textbf{Mean (\%)} & \textbf{Std Dev (pp)} \\
\midrule
Llama 3.2 3B & 24 & 36.4 & 95.4 & 59 & 80.8 & 19.0 \\
Llama 3.1 70B & 24 & 71.5 & 93.2 & 21.7 & 83.1 & 6.1 \\
Qwen 2.5 72B & 8 & 71.5 & 93.0 & 21.2 & 80.7 & 8.4 \\
\bottomrule
\end{tabular}
\end{table}

\begin{figure}[!htbp]
\centering
\includegraphics[width=0.95\textwidth]{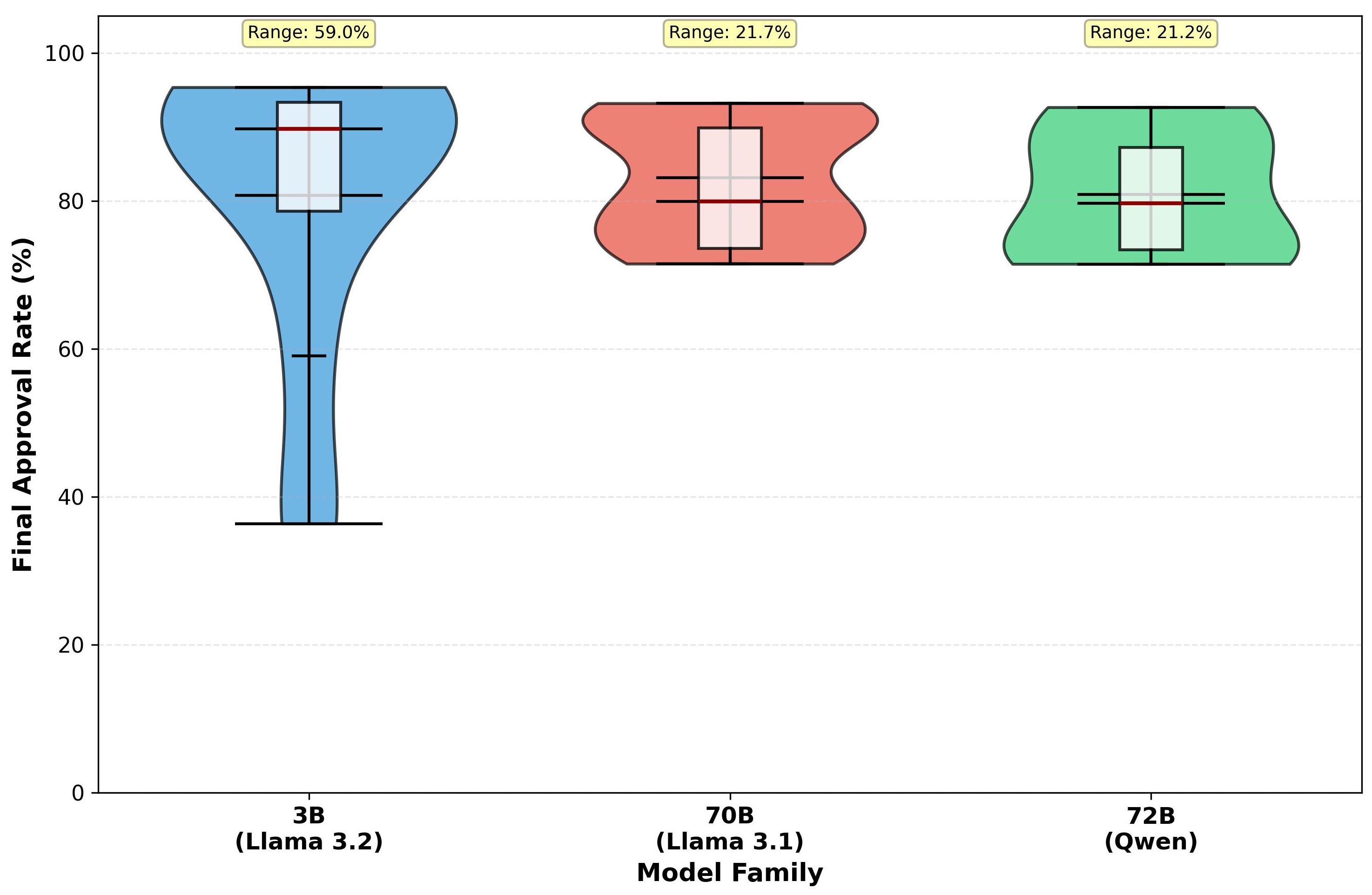}
\caption{Distribution of approval rates across sequential orderings by model family. The violin plots reveal that ordering instability is not a minor perturbation but a first-order determinant of system behavior. Llama 3B exhibits a bimodal distribution spanning a 59.0 percentage point range, with the majority of orderings clustered in two regions: high approval ($\sim$85--95\%) and moderate approval ($\sim$60--80\%), depending primarily on which role appears first. Larger models (70B, 72B) show tighter, more unimodal distributions centered around 80\%, but still exhibit 21+ percentage point ranges. The narrower shape of the larger model distributions reflects stronger consensus within orderings, combined with persistent sensitivity to which ordering is selected.}
\label{fig:approval_distribution}
\end{figure}

Figure~\ref{fig:approval_distribution} visualizes these distributions using violin plots, which show both the range and the density of approval rates across orderings. The width of each violin at a given approval rate indicates how many orderings produced that rate. The box plots embedded within each violin show the median (center line), interquartile range (box), and full range (whiskers).

These results reveal three critical patterns:

\paragraph{Substantial sensitivity in small models.} Llama 3B exhibits a 59.0 percentage point range visible in the violin plot, with approval rates varying from approximately 37\% to 96\% depending solely on agent ordering. This represents a substantial qualitative regime shift induced by topology alone. The bimodal shape of the 3B distribution indicates two distinct behavioral clusters: permissive orderings (Consumer Advocate or Data Scientist first) that cluster around 85--95\% approval, and conservative orderings (Risk Manager or Regulator first) that cluster around 60--80\% approval.

\paragraph{Persistent sensitivity at scale.} Both 70B-parameter models retain substantial ordering effects despite their enhanced reasoning capabilities. The 21.7 percentage point range for Llama 70B and 21.2 percentage point range for Qwen 72B demonstrate that scaling does not eliminate topological sensitivity. Larger models do not ``reason their way out'' of path dependence; instead, they exhibit tighter consensus (narrower violin shapes, lower variance within orderings) combined with persistent sensitivity to which agent sets the initial frame. The reduced spread within orderings reflects stronger inter-agent coupling, not improved robustness to architectural choices.

\paragraph{Consistency across model families.} The fact that three independently trained model families (Llama and Qwen, spanning 3B to 72B parameters) all exhibit strong ordering effects rules out explanations based on idiosyncratic training artifacts or architecture-specific quirks. Ordering instability is a fundamental property of sequential multi-agent interaction, not a correctable defect of particular models. The persistent 21+ percentage point ranges at 70B and 72B scales demonstrate that this is a structural phenomenon inherent to sequential topology, not a capability limitation that disappears with better models.

\subsection{Role position effects}
\label{app:ordering_results:position}

Examination of the heatmaps reveals systematic patterns in which roles drive high or low approval rates when placed in early positions. In Llama 3B (Figure~\ref{fig:llama3b_heatmap}), orderings with Consumer Advocate or Data Scientist in the first position consistently produce higher approval rates (rightmost columns, lighter colors), while orderings with Risk Manager or Regulator first produce lower approval rates (leftmost columns, darker colors). This pattern reflects the different epistemic priors and evaluative frames each role brings: Consumer Advocate emphasizes access and inclusion, while Risk Manager emphasizes downside protection.

Crucially, these are not differences in inherent role bias measured in isolation. When roles act as single agents (baseline condition, not shown in heatmaps), their individual approval rates are much closer together. The divergence emerges from interaction: the first agent sets an interpretive frame and confidence level that subsequent agents condition on, amplify, or fail to challenge. In large models (Figures~\ref{fig:llama70b_heatmap} and \ref{fig:qwen72b_heatmap}), this amplification is stronger, producing the more uniform within-column coloring that indicates tighter consensus once an initial direction is established.

\subsection{Implications for safety evaluation}
\label{app:ordering_results:implications}

These results invalidate the assumption that multi-agent safety can be evaluated by testing a single fixed architecture. A system that passes safety benchmarks under one agent ordering may produce drastically different outcomes under permutation. For Llama 3B, this is not a minor perturbation but a qualitative behavioral shift: the difference between a 59 percentage point range in approval rates represents a fundamental change in system function, not a boundary case or edge condition.

Current evaluation protocols test models on fixed tasks and report single-number accuracies or fairness metrics. Our results demonstrate that for sequential multi-agent systems, there is no single accuracy or fairness number to report. Instead, there is a distribution over orderings, and the width of that distribution is a first-order safety property. A model with high accuracy under researcher-selected ordering but wide variance under permutation is not safe for deployment.

The consistency of these effects across model scales further undermines the assumption that future, larger models will naturally ``solve'' coordination problems through better reasoning. Scaling improves local capabilities while simultaneously increasing global coupling, making early decisions more influential and subsequent corrections less likely. This represents a scale paradox in agentic AI safety: the very properties that make individual models more capable (coherence, confidence calibration, consistency) make multi-agent systems more brittle and topology-dependent.
\section{Parallel Topology and Functional Collapse}
\label{app:parallel}

In parallel topology, all four roles evaluate each application independently without observing one another's outputs. A judge agent then aggregates their recommendations into a single system-level decision. While this design removes the ordering instability and cascade dynamics inherent to sequential topology, it introduces a distinct topology-driven failure mode: functional collapse through approval saturation.

\subsection{Overall approval rates by architecture}
\label{app:parallel:rates}

Table~\ref{tab:parallel_approval} reports overall approval rates for parallel aggregation compared to sequential deliberation (mean across orderings). Results demonstrate that parallel topology fundamentally alters system behavior relative to sequential topology, with the magnitude of the effect varying by model scale.

\begin{table}[!htbp]
\centering
\caption{Approval rates across architectures. Parallel topology produces substantially higher approval rates in small models and comparable rates in large models, demonstrating that topology determines the dominant failure mode rather than eliminating safety concerns.}
\label{tab:parallel_approval}
\begin{tabular}{lccc}
\toprule
\textbf{Model} & \textbf{Sequential Mean (\%)} & \textbf{Parallel (\%)} & \textbf{Difference (pp)} \\
\midrule
Llama 3.2 3B & 81.4 & 98.3 & +16.9 \\
Llama 3.1 70B & 83.1 & 81.7 & $-$1.4 \\
\bottomrule
\end{tabular}
\end{table}

For Llama 3B, the shift from sequential to parallel topology increases approval rates by nearly 17 percentage points. For Llama 70B, the topologies produce comparable overall rates. However, as Section D.2 demonstrates, these aggregate statistics conceal critical differences in how the systems discriminate risk. The key finding is not that scale solves topology problems, but that \textit{different topologies produce different failure modes}, and neither topology is inherently safe.

\subsection{Functional collapse: parallel topology abandons risk discrimination}
\label{app:parallel:discrimination}

\begin{figure}[!htbp]
\centering
\includegraphics[width=\textwidth]{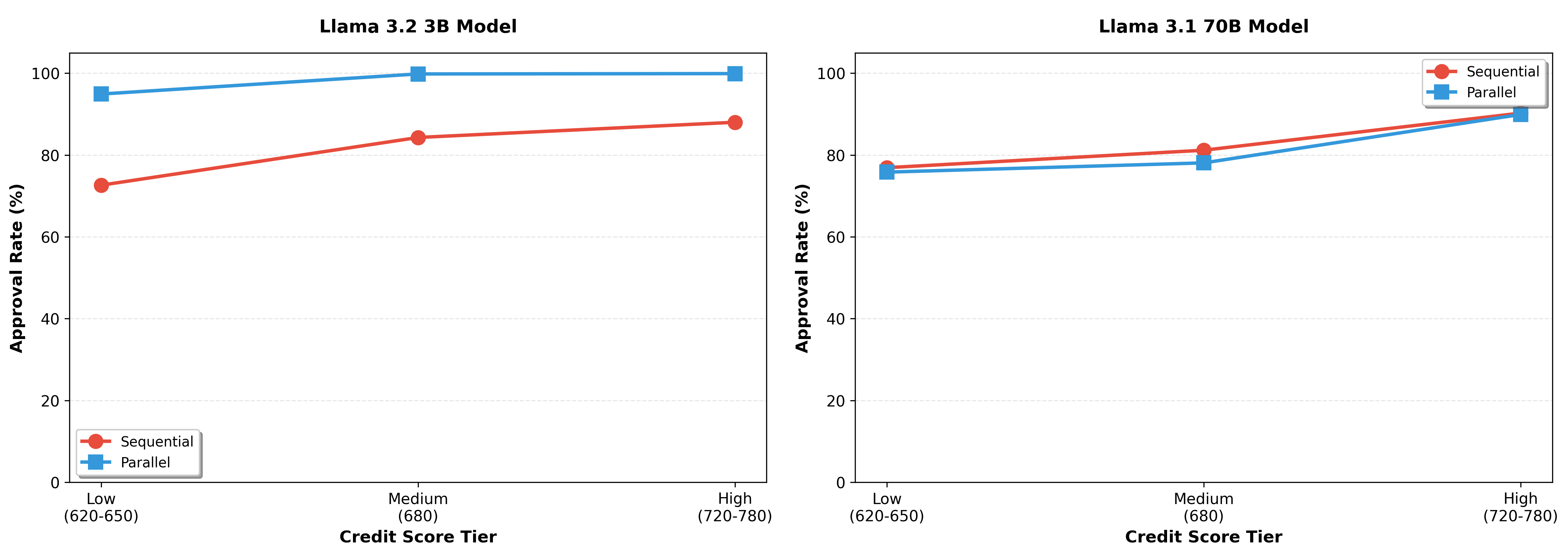}
\caption{Approval rates by credit score tier under sequential and parallel topologies. Left panel: Llama 3B parallel topology exhibits functional collapse, with a nearly flat approval curve (95--100\%) that abandons risk discrimination. Sequential topology maintains a 15 percentage point spread across credit tiers. Right panel: Llama 70B parallel topology maintains discrimination, but both topologies exhibit their characteristic failure modes—parallel produces uniform high approval, sequential exhibits ordering instability (see Appendix C). The critical observation is that topology choice determines which pathology dominates, not whether the system is safe.}
\label{fig:parallel_by_credit}
\end{figure}

Figure~\ref{fig:parallel_by_credit} stratifies approval rates by credit score tier, revealing how parallel topology alters the relationship between applicant risk and system decisions. The comparison demonstrates that parallel aggregation introduces a distinct pathology: the system satisfies aggregate fairness metrics while losing the capacity to perform its core function.

\paragraph{Llama 3B: Topology determines task function.} In sequential topology, the 3B model maintains a 15 percentage point spread across credit tiers (73\% for low-credit, 88\% for high-credit), demonstrating sensitivity to risk signals despite suffering from severe ordering instability (Appendix C shows a 59 percentage point range across orderings). In parallel topology, the system approves 95--100\% of applicants regardless of credit quality, producing a nearly flat curve with only a 5 percentage point spread. This represents functional collapse: the system satisfies demographic parity by approving nearly everyone, but provides zero value for credit assessment.

The parallel topology does not \textit{improve} the sequential topology's ordering instability; it \textit{trades} ordering instability for approval saturation. Sequential topology produces outcomes that vary with agent order but maintain risk sensitivity within each ordering. Parallel topology produces stable outcomes that are insensitive to ordering but abandon the task objective. \textbf{Neither topology is safe they fail in different ways.}

\subsection{Mechanism: topology-induced bottleneck creates systematic bias}
\label{app:parallel:mechanism}

The approval saturation observed in parallel 3B topology arises from the architectural structure of parallel aggregation, not from individual agent defects. When four independent evaluators produce conflicting recommendations, the judge must consolidate their outputs without access to the reasoning chains that would emerge from sequential interaction. This creates a centralized decision bottleneck: all uncertainty is funneled through a single aggregation step.

In small models, this bottleneck induces systematic approval bias. The judge, faced with mixed signals (e.g., 2 approve, 2 deny), defaults toward permissive decisions rather than engaging in risk-sensitive adjudication. This is not a correctable defect but an emergent property of the topology: parallel structure concentrates decision authority in the judge while removing the contextual information (prior agent reasoning) that could inform balanced judgment.

In large models, the judge can perform this aggregation without systematic bias, but the parallel topology still eliminates the reasoning diversity that sequential interaction enables. The tradeoff is inherent to the architectural choice: parallel topology removes cascades by removing inter-agent influence, but this same removal prevents collaborative refinement of judgments.

\subsection{Implications: no topology is universally safe}
\label{app:parallel:implications}

These results demonstrate the core thesis of this paper: \textit{interaction topology determines system behavior in ways that cannot be reduced to component-level properties}. The same agents, prompts, and task produce radically different outcomes under sequential versus parallel topology. Neither topology is inherently safe:

\begin{itemize}
\item \textbf{Sequential topology} suffers from ordering instability (Appendix C: 59pp range for 3B, 21pp range for 70B) and information cascades (Section 3.2: 99.9\% agreement in 70B with zero error correction). These are structural properties of sequential interaction, not correctable through better prompts or alignment.

\item \textbf{Parallel topology} suffers from functional collapse in small models (this appendix: 95--100\% approval regardless of risk) and eliminates deliberative reasoning in large models (judge aggregates independent votes without collaborative refinement). These are structural properties of parallel aggregation, not correctable through better voting rules.
\end{itemize}

The implication is not that scale solves topology problems, but that \textit{topology choice determines which failure mode dominates}. Deployment decisions must account for these architectural tradeoffs rather than assuming that any single topology is optimal. Current safety evaluation, which tests models under researcher-selected architectures, systematically fails to characterize topology-driven pathologies. A system that appears safe under parallel aggregation may exhibit severe ordering instability under sequential deliberation, and vice versa.

Topology-aware evaluation requires stress-testing across architectural variations. For sequential systems, this means bounding variance across orderings and measuring cascade strength. For parallel systems, this means verifying that aggregation preserves task-relevant discrimination rather than satisfying fairness metrics through universal approval. Neither form of robustness can be assumed from component-level testing, and neither topology eliminates the need for architectural validation.

\FloatBarrier

%%%%%%%%%%%%%%%%%%%%%%%%%%%%%%%%%%%%%%%%%%%%%%%%%%%%%%%%%%%%%%%%%%%%%%%%%%%%%%%
%%%%%%%%%%%%%%%%%%%%%%%%%%%%%%%%%%%%%%%%%%%%%%%%%%%%%%%%%%%%%%%%%%%%%%%%%%%%%%%

\end{document}